\pdfoutput=1

\documentclass[10pt,a4paper]{article}

\usepackage[final]{lrec2026}

\usepackage[T1]{fontenc}
\usepackage[utf8]{inputenc}
\usepackage{microtype}
\usepackage{amsmath}
\usepackage{amssymb}
\usepackage{booktabs}
\usepackage{multirow}
\usepackage{latexsym}
\usepackage{graphicx}
\usepackage{hyperref}
\usepackage{float}
\usepackage{comment}
\usepackage{url}
\title{J-CHAT: Japanese Large-scale Spoken Dialogue Corpus for Spoken Dialogue Language Modeling}

\name{Wataru Nakata$^{1*}$, Kentaro Seki$^{1*}$, Hitomi Yanaka$^{1}$\\{\bf\large Yuki Saito$^{1}$ Shinnosuke Takamichi$^{1,2}$, Hiroshi Saruwatari$^{1}$}}

\address{$^1$The University of Tokyo, Japan\\
$^2$Keio University, Japan\\
\texttt{nakata-wataru855@g.ecc.u-tokyo.ac.jp}\\
$^*$Equal contribution.}

\abstract{
Spoken dialogue is essential for human-AI interactions, providing expressive capabilities beyond text. Developing
effective spoken dialogue systems (SDSs) requires large-scale, high-quality, and diverse spoken dialogue corpora.
However, existing datasets are often limited in size, spontaneity, or linguistic coherence. To address these limitations, we
introduce J-CHAT, a 76,000-hour open-source Japanese spoken dialogue corpus. Constructed using an automated,
language-independent methodology, J-CHAT ensures acoustic cleanliness, diversity, and natural spontaneity. The corpus is built from YouTube and podcast data, with extensive filtering and denoising to enhance quality. Experimental
results with generative spoken dialogue language models trained on J-CHAT demonstrate its effectiveness for SDS
development. By providing a robust foundation for training advanced dialogue models, we anticipate that J-CHAT will
drive progress in human-AI dialogue research and applications.
\\ \newline \Keywords{spoken dialogue corpus, speech datasets, Japanese, spoken language modeling, language resources}}

\begin{document}
\maketitleabstract

\section{Introduction}
To realize human-AI interaction through spoken language, the development of spoken dialogue systems (SDSs) is crucial.
Traditional SDSs have been constructed by cascading multiple modules, such as speech recognition, response generation,
and speech synthesis~\cite{huang2024audiogpt}. However, this cascaded SDS approach struggles to account for
subtle nuances and nonverbal vocal expressions, which are often lost in transcription yet are critical elements that distinguish spoken dialogue from text-based chat. Recently, it has been demonstrated that end-to-end SDSs can be realized using machine learning methods trained on large-scale data, and several such methods have been proposed~\cite{nguyen2023generative,mitsui2023towards}. This approach is gaining attention because it is expected to enable smoother human-AI interaction by
fully leveraging the information conveyed through spoken language.

The realization of end-to-end SDS requires large-scale datasets.  
For instance, previous studies on synthesizing dialogue speech using end-to-end SDS have utilized 20k hours of speech~\cite{nguyen2023generative} and 100k hours of speech~\cite{borsos2023soundstorm}, suggesting that tens of thousands of hours of dialogue speech corpora are necessary to achieve this approach.  
In addition to size, the quality and diversity of the corpus are also crucial.  
In dialogue systems, since the final output is speech, it is desirable for the training data to be clean and free of noise.  
Furthermore, machine learning models tend to degrade in performance when applied to out-of-domain data.  
Therefore, it is desirable to have a diverse dataset, particularly including spontaneous speech, which is often not covered by existing studio-recorded corpora.  

However, large-scale open-source dialogue speech corpora remain scarce, and methods for their construction are underdeveloped. 
The largest available resource, Seamless Interaction~\cite{agrawal2025seamless}, contains over 4,000 hours of dialogue speech. 
While this corpus is sufficiently large for training SDSs, its manual recording under controlled conditions makes it financially impractical to reproduce in other languages. 
Although several studies have investigated automatic methods for constructing large-scale speech corpora in recent years, these efforts have primarily targeted speech recognition~\cite{chen2021gigaspeech, takamichi2021jtubespeech} or speech synthesis~\cite{seki2023text}, typically segmenting speech into short utterance-level units.  
Consequently, such corpora lack key characteristics of dialogue, including linguistic coherence across turns and natural turn-taking dynamics. 
To date, no general method for automatically constructing a large-scale dialogue speech corpus has been explored.

To address this issue, we propose a method for constructing large-scale dialogue speech corpora and have publicly released the constructed corpus, named ``Japanese Corpus for Human-AI Talks'' (J-CHAT)\footnote{\url{https://huggingface.co/datasets/sarulab-speech/J-CHAT}\label{corpus_url}}.  The corpus is distributed under CC BY-NC 4.0 license for ``information analysis'' which is defined in Japanese copyright act article 30-4\footnote{\url{https://laws.e-gov.go.jp/law/345AC0000000048\#Mp-Ch_2-Se_3-Ss_5-At_30_4}\label{copyright}}.
Our corpus construction method automatically filters and collects data on a dialogue-unit basis from the internet.  
Since the entire process is automated, it is highly scalable.  
The constructed J-CHAT corpus contains 76k hours of Japanese dialogue speech data, which is comparable to the existing large-scale corpora for the development of end-to-end SDSs.  
Additionally, background music removal procedures ensure the acoustic cleanliness of the dataset.  
Moreover, by collecting wild data from multiple domains, the corpus includes spontaneous speech and consists of diverse data.

Our contributions are as follows:  
\begin{itemize} 
\item We propose a method for constructing a large-scale corpus for end-to-end SDSs. Our proposed method constructs a corpus that is acoustically clean, diverse, and includes spontaneous speech.
\item Our proposed method is an automated and language-independent approach, making it easy to construct corpora for other languages.  
\item We constructed and released the Japanese dialogue speech corpus, J-CHAT.  Also, we experimentaly validated its effectiveness for end-to-end SDSs.
\end{itemize}

\newcommand{\doublecirc}{{\ooalign{$\checkmark$\crcr\hss$\circ$\hss}}}
\newcommand{\Yes}{$\checkmark$}
\newcommand{\No}{~}
\newcommand{\None}{~}

\newcommand{\singleline}[6]{#1 & #2 & #3 & #4 & #6 & #5 \\}

\begin{table*}
\centering
\caption{
Comparison of speech corpora related to this study, sorted by size.
}
\label{table:corpus_compare}
\scalebox{0.8}{
\begin{tabular}{lccccc}
\hline
\singleline{\textbf{Corpus name}}{\textbf{Size(hours)}}{\textbf{Open-source}}{\textbf{Dialogue}}{\textbf{Clean speech}}{\textbf{Spontaneous}}
\hline
\singleline{STUDIES~\cite{saito2022studies}}{$8.2$}{$\checkmark$}{$\checkmark$}{$\checkmark$}{\No}
\singleline{DailyTalk~\cite{Lee2023dailytalk}}{$20$}{$\checkmark$}{$\checkmark$}{$\checkmark$}{$\checkmark$}
\singleline{CallHome-JP~\cite{callhome}}{$49$}{$\checkmark$}{$\checkmark$}{$\checkmark$}{$\checkmark$}
\singleline{MultiDialog~\cite{park-etal-2024-lets}}{$340$}{$\checkmark$}{$\checkmark$}{\No}{$\checkmark$}
\singleline{LibriTTS~\cite{zen2019libritts}}{$585$}{$\checkmark$}{\No}{$\checkmark$}{\None}
\singleline{SSSD~\cite{sheikh2025scalable}}{$727$}{$\checkmark$}{$\checkmark$}{$\checkmark$}{\No}
\singleline{Fisher~\cite{cieri2004fisher}}{$2k$}{\No}{$\checkmark$}{$\checkmark$}{$\checkmark$}
\singleline{Seamless Interaction~\cite{agrawal2025seamless}}{$4k$}{$\checkmark$}{$\checkmark$}{$\checkmark$}{$\checkmark$}
\singleline{GigaSpeech~\cite{chen2021gigaspeech}}{$33k$}{$\checkmark$}{\No}{\No}{$\checkmark$}
\singleline{\textbf{J-CHAT~(This study)}}{$76k$}{$\checkmark$}{$\checkmark$}{$\checkmark$}{$\checkmark$}
\hline
\end{tabular}
}
\end{table*}

\begin{figure*}[t]
    \centering
    \includegraphics[width=\linewidth]{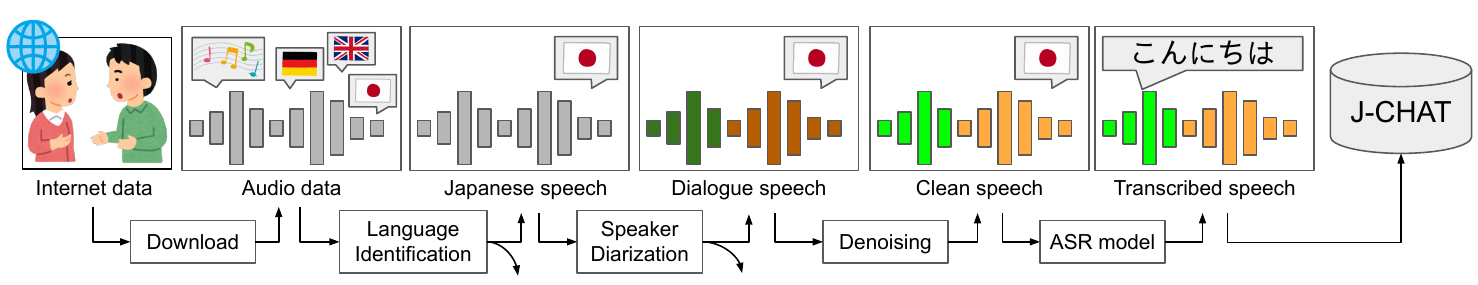}
    \caption{
        Corpus construction methodology proposed in this work.
    }
    \label{fig:corpus_construction}
\end{figure*}

\section{Related Work}

A comparison between existing corpora and our
J-CHAT corpus is shown in Table~\ref{table:corpus_compare}. The STUDIES corpus~\cite{saito2022studies} is an open-source
conversational speech corpus; however, since it
consists of scripted speech recorded in a studio, it
does not include spontaneous speech and covers only limited domains. 
The DailyTalk corpus (Lee et al., 2023) is also an open-source conversational speech corpus, but it contains only about 20 hours of data, which is relatively small in scale. 
The CallHome Japanese subset (hereinafter referred to as “CallHome-JP”) is
the largest spontaneous conversational speech
dataset in Japanese. Although it is open-source, its scale remains insufficient for training end-to-end SDSs. 
The MultiDialog corpus~\cite{park-etal-2024-lets}, and the SSSD~\cite{sheikh2025scalable} corpus are English spontaneous conversational speech datasets, but their size are still insufficient for end-to-end SDS.
The Fisher corpus~\cite{cieri2004fisher} is a sufficiently large conversational speech dataset, but it is not open-source, making experimental reproducibility challenging.  
The LibriTTS corpus~\cite{zen2019libritts} and GigaSpeech corpus~\cite{chen2021gigaspeech} are open-source corpora; however, as it is designed for speech synthesis and  recognition, respectively, its data is segmented at the utterance level, lacking conversational context-specific information. 
Recently, Seamless Interaction~\cite{agrawal2025seamless} has been released which is over 4k hours in duration making it a viable training dataset for SDSs. However, their corpus construction methodology was based on manual recording in a controlled environment limiting the scalability required to reproduce SDSs in other languages.

In contrast, our J-CHAT corpus was collected from in-the-wild sources, allowing for the inclusion of diverse speech. Furthermore, using in-the-wild sources allows scalability of the dataset size making the construction SDSs in many languages possible. 
With a scale of 76k hours, which is over 1,000 times larger than CallHome-JP, it provides a dataset large enough for the development of end-to-end SDSs.
Furthermore, as an open-source resource, it is expected to facilitate research and development in SDSs.  
Unlike utterance-level segmentation, J-CHAT is segmented at the dialogue level, which enables modeling of discourse phenomena such as contextual coherence and turn-taking behavior.

\section{Corpus Construction Methodology}

Figure 1 illustrates the overall workflow of our corpus construction process. To build a spoken dialogue corpus for a target language (Japanese in this study), we collected audio data from the internet. We then removed inappropriate content by performing language identification, dialogue extraction, and background-noise removal. Finally, we transcribed the spoken content using an automatic speech recognition (ASR) model.

\subsection{Data Collection}
Since previous research~\cite{takamichi2021jtubespeech} has demonstrated that a diverse range of speech can be collected from YouTube, we used it as one of our primary data sources. We searched YouTube using randomly selected Wikipedia page titles as keywords, following previous research~\citep{takamichi2021jtubespeech}, which resulted in approximately 600k audio files totaling about 180k hours of speech.

YouTube contains not only dialogue-dominant videos but also non-speech videos like music and monologue videos like game commentaries.
This resulted in a low proportion of dialogue data, making it challenging to secure a sufficient amount of data.
To address this, we also collected data from podcasts to expand the scale of our dataset. 
Podcasts are speech platforms, making it efficient to gather speech data from them.
Furthermore, PodcastIndex\footnote{\url{https://podcastindex.org/}} provides extensive metadata, including labels that indicate which language the content is.
We retrieved RSS feed URLs of all podcast stations labeled as Japanese from PodcastIndex. 
Subsequently, we searched and downloaded for any audio URLs listed on the collected RSS feeds. 
As a result, we obtained approximately 880k audio files totaling around 140k hours of audio data.

\subsection{Data Selection}
\subsubsection{Extracting Japanese Speech Data}
%
To filter out non-Japanese audio, we used Whisper’s language-identification model (Radford et al., 2023) and retained segments only when the probability of Japanese speech (\textit{p}) exceeded 0.8. This process retained $55.7\%$ of the YouTube data and $84.7\%$ of the podcast data.

\subsubsection{Extracting Dialogue Speech Data}
From this Japanese speech dataset, we specifically extracted dialogue segments. This requires detecting conversation regions within the audio. Speaker diarization (SD) techniques are commonly used to identify who is speaking and when.
SD analyzes the audio data to detect speech segments and links segments spoken by the same speaker, outputting pairs of speech segments and speaker IDs.
In this study, we used a publicly available pre-trained SD model, PyAnnote~\cite{Plaquet23}, to obtain speech segments with speaker IDs.

Based on the speech segments (hereafter referred to as turns), we split the audio data into separate dialogues at gaps of 5 seconds or more.
Next, we filtered out dialogues where a single speaker's turns account for more than $80\%$ of the time, ensuring that only valid conversations are selected.
This treats dialogues dominated by a single speaker as monologues.
Here, we allowed dialogues with two or more distinct speaker IDs as valid types of conversations.

Through the above process, we obtained pairs of dialogue-speech data and their labels (each turn’s duration and speaker ID).
The proportion of Japanese speech data containing dialogues was $41.9\%$ for the YouTube data and approximately $45.0\%$ for the podcast data.

\subsection{Data Cleansing}
YouTube and podcasts often include background music (BGM), which acts as noise for speech generation models, so it needs to be removed. 
Techniques for extracting speech from data mixed with BGM are studied in the fields of speech enhancement and source separation, and recently, machine learning models have achieved high performance in this area~\cite{koizumi2023libritts,rouard2022hybrid}. 
We applied a pre-trained speech enhancement model~Demucs~\cite{rouard2022hybrid} for data cleansing to all the collected audio and obtained the J-CHAT corpus.

\subsection{ASR}
Some SDSs require transcription of the speech for training~\cite{kyutai2024moshi}. To meet this need, we created the transcription using ASR model.
For ASR model, we used reazonspeech-nemo-v2\footnote{\url{https://huggingface.co/reazon-research/reazonspeech-nemo-v2}}.
For each subword in transcript, the alignment information is also provided.

\section{Corpus Analysis}
\label{sec:corpus_analysis}

\subsection{Dataset Size}
\label{dataset size}
\begin{table}[t]
    \centering
    \caption{Corpus statistics by its subsets, YouTube and Podcast. \# means ``number of''.}
    \label{tab:analysis_table}
    \scalebox{0.8}{
    \begin{tabular}{r|r|r|r}
    \toprule
        feature & YouTube & Podcast & Total  \\\midrule
        total duration[hr] & 11,017 & 65,019 & 76,036\\
        \# dialogue & 1,015,109 & 4,409,405 & 5,424,514 \\ 
        mean duration [s] & 39.07 &  53.11 & 50.23\\ 
        mean \# turns & 7.58 & 10.68 & 10.10\\ 
        mean \# speakers &3.23 &3.12&3.14\\ \bottomrule
    \end{tabular}
    }
\end{table}

Table~\ref{tab:analysis_table} shows the statistical analysis of J-CHAT corpus.
J-CHAT consists of a total of 76k hours of Japanese speech data, with the YouTube subset accounting for 11k hours and the podcast subset for 65k hours.
The corpus also provides predefined train/valid/test/other splits for each YouTube and Podcast subsets for the reproducibility of research.
For the YouTube subset, the durations of the train/valid/test/other splits are 10872.5/108.7/1.2 hours respectively.
For the Podcast subset the duration of train/valid/test/other splits are 57291.5/575/1.3 hours respectively.

A notable difference between the subsets is that the average duration of dialogues in the podcast subset is approximately 1.4 times longer, which is attributed to a higher number of turns per dialogue.
On the other hand, the average number of speakers per dialogue is nearly the same for both subsets.

\subsection{Acoustic Cleanliness}

To ensure that the J-CHAT corpus is not significantly affected by noise, we evaluated its acoustic cleanliness using NISQA~\cite{mittag2021nisqa} and compared it with CallHome-JP (telephone recordings) and STUDIES (studio recordings).

The average quality scores were as follows: $4.01$ for STUDIES, $1.98$ for CallHome-JP, $2.37$ for the YouTube subset of J-CHAT, and $2.99$ for the Podcast subset of J-CHAT. Although there is a quality gap between J-CHAT and the studio-recorded STUDIES corpus, J-CHAT’s quality is superior to that of CallHome-JP, the largest spontaneous dialogue corpus for Japanese. Furthermore, the Podcast subset of J-CHAT achieved a higher score than the YouTube subset, indicating the effectiveness of using podcasts as a data source for constructing dialogue corpora.
This suggests that the noise level in the Podcast subset is sufficiently low for use in speech synthesis.  

\subsection{Dataset Diversity}
\begin{figure}[t]
    \centering
    \includegraphics[width=\linewidth]{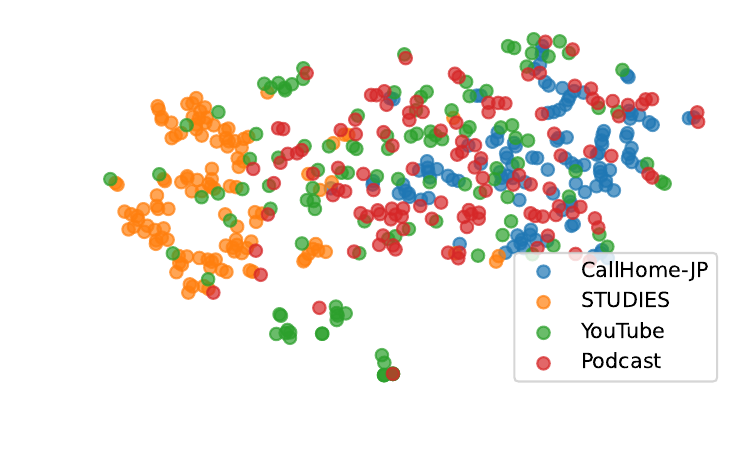}
    \caption{t-SNE visualization of sentence embeddings extracted from the ASRed transcripts from Podcast and YouTube subset of J-CHAT, STUDIES~\cite{saito2022studies} and CallHome-JP~\cite{callhome} corpus. For each corpus, random 120 dialouges are extracted for calculation. }
    \label{fig:tsne-linguistic}
\end{figure}
{\bf Linguistic Diversity:}
As a dialogue corpus, J-CHAT is expected to cover a wide range of topics.
To analyze its topic diversity, we compared the distribution of sentence embeddings from J-CHAT transcriptions with those from STUDIES (scripted) and CallHome-JP (spontaneous).  
We randomly selected $120$ dialogues from the YouTube subset of J-CHAT, and the Podcast subset of J-CHAT as well as the all dialogues in STUDIES. For CallHome-JP, we extracted all $120$ dialogues. Subsequently sentence embeddings are extracted using a pre-trained sentence embedding model\footnote{\url{https://huggingface.co/sonoisa/sentence-bert-base-ja-mean-tokens-v2}}.  

The t-SNE plot of the embeddings is shown in Figure~\ref{fig:tsne-linguistic}.  
The results indicate that each subset of J-CHAT exhibits a broader distribution compared to CallHome-JP.  
Note that, although the figure based on this comparison at the same scale may make J-CHAT appear sparse, in reality, J-CHAT is significantly larger than CallHome-JP, ensuring that the region covered by CallHome-JP also contains a sufficient amount of data.  
Additionally, the distribution of STUDIES is distinct from those of CallHome-JP and J-CHAT because the dialogue topics of STUDIES are limited in conversations between a teacher and students in school.

Furthermore, to conduct a quantitative analysis, we calculated the average pairwise cosine similarity for each dataset.  
The values for CallHome-JP, STUDIES, the YouTube subset, and the Podcast subset are $0.6164$, $0.5186$, $0.2390$, and $0.3457$, respectively.  
Since a lower cosine similarity indicates lower similarity, these results quantitatively demonstrate that the J-CHAT corpus covers a more diverse range of topics.  

\begin{figure}[t]
    \centering
    \includegraphics[width=0.9\linewidth]{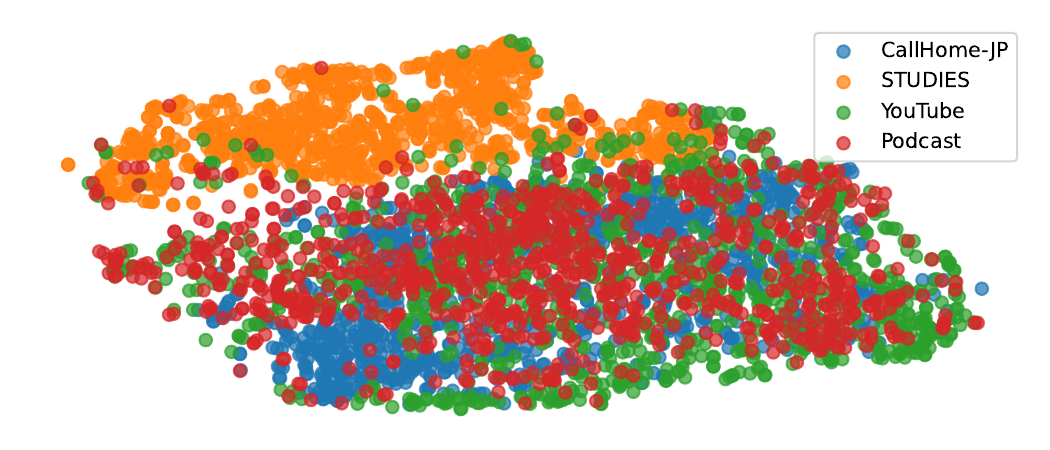}
    \caption{
        Distribution of HuBERT~\cite{hsu2021hubert} features extracted from J-CHAT (ours), STUDIES (simulated dialogue), and CallHome-JP (real dialogue).
    }
    \label{fig:acoustic_diversity}
\end{figure}
\hfill\break
\noindent {\bf Phonetic Diversity:}
To demonstrate that J-CHAT includes spontaneous speech, we conducted an analysis of acoustic features to confirm that J-CHAT contains expression characteristics of spontaneous speech.  
In this analysis, we used HuBERT~\cite{hsu2021hubert} features, which are expected to capture phonetic information, and compared them with the STUDIES corpus, which consists of scripted dialogue speech, and CallHome-JP, which consists of spontaneous dialogue speech.  

We randomly sampled 1,000 frame-wise HuBERT features from each subset of J-CHAT, as well as from STUDIES~\cite{saito2022studies} and CallHome-JP~\cite{callhome}. 
For each sampled point, we computed HuBERT features by randomly selecting 5-second intervals.

The t-SNE plot is shown in Figure~\ref{fig:acoustic_diversity}.  
The distributions for STUDIES were found to be limited to specific regions.  
In contrast, the subsets of J-CHAT and CallHome-JP encompassed both regions.  
These results indicate that J-CHAT covers phonetic diversity in spontaneous dialogue speech. 

\section{Experiments}
To validate that J-CHAT corpus is a large-scale corpus suitable for training dialogue-oriented spoken language models, we trained and evaluated the performance of dialogue-generative spoken language model (dGSLM,~\cite{nguyen2023generative}).
dGSLM follows a framework proposed by~\cite{lakhotia2021generative} which consists of three distinct modules: speech-to-unit, unit language model, and unit-to-speech (vocoder). 

Our experiments include three models and resynthesis samples for comparison: a resynthesized J-CHAT test subset utilizing speech-to-unit and vocoder (resynth), dGSLM on the YouTube subset of J-CHAT (dGSLM-YouTube), dGSLM trained on the podcast subset (dGSLM-podcast), and dGSLM trained on all subsets of J-CHAT (dGSLM-J-CHAT).
Trained models, generated samples and training data are available\footref{corpus_url}.

\subsection{Experimental Conditions}

We split the dialogues in the J-CHAT corpus into two channels by swapping the output channel of speech according to the turn-taking event. We used the train/valid/test sets explained in Section~\ref{dataset size}.
Then, we performed discretization of speech using k-means clustering on HuBERT-extracted features using train set from both YouTube and Podcast subsets. The number of clusters for k-means was set to 1,000.

For the vocoder used for speech generation from the discretized speech, we used HiFi-GAN~\cite{NEURIPS2020_c5d73680} conditioned with speaker information from XVector~\cite{snyder2018xvector}, as used in previous work~\cite{kharitonov-etal-2022-text}. 

\subsection{Training Details}
For the implementation of dGSLM, we used official implementation of the model\footnote{\url{https://github.com/facebookresearch/fairseq}}.
For HuBERT model used in the speech-to-unit, we used the pretrained Japanese HuBERT model~\cite{sawada2024release}\footnote{\url{https://huggingface.co/rinna/japanese-hubert-base}}.
For k-means clustering, we used the original implementation of the dGSLM except for the number of clusters which was set to $1,000$.
The dGSLM model was configured as in the previous research~\cite{nguyen2023generative}.

For training dGSLM model, we used 32 NVIDIA V100 GPUs with the learning rate of $2\times 10^{-4}$. The training was performed for $100,000$ steps.
The batch size was $36,864$ tokens and the the maximum number of tokens per sample are set to $3,000$.
This was equivalent to the $1$ minute of dialogue.
For other hyper-parameters regarding the training, we followed the original paper~\cite{nguyen2023generative}.

\begin{table*}[t]
    \centering
    \caption{MOS test results with their 95\% confidence intervals.}
    \label{tab:result}
    \begin{tabular}{l|rr}
\toprule
Model &  Naturalness & Meaningfulness \\\midrule
resynth         & 2.55 $\pm$ 0.18& 2.48 $\pm$ 0.18\\\midrule
dGSLM-Youtube   & 1.44 $\pm$ 0.13& 1.56 $\pm$ 0.14\\
dGSLM-podcast   & 1.44 $\pm$ 0.13& 1.52 $\pm$ 0.13\\
dGSLM-J-CHAT    & 2.28 $\pm$ 0.19& 2.18 $\pm$ 0.19\\
\bottomrule
\end{tabular}
\end{table*}

For the XVector conditioned vocoder, we used the pretrained XVector model\footnote{\url{https://huggingface.co/speechbrain/spkrec-xvect-voxceleb}}.
The vocoder was trained with the JVS~\cite{takamichi2020jsut} and JVNV~\cite{xin2024jvnv} corpus.
These two corpora include Japanese reading-style, studio-quality speech without/with non-verbal expression, respectively.
Training of vocoder took 2 days with 4 NVIDIA V100 GPUs.

\subsection{Evaluation}\label{sec:evaluation}

For evaluation, we performed subjective listening tests on the mean opinion score (MOS) on the naturalness and meaningfulness of dialogues following previous work~\cite{nguyen2023generative}.
For each subjective listening test, we recruited 60 Japanese native speakers, and each listener evaluated 8 samples.
For sample generation used in the evaluation, we used the first 5 seconds of the test set derived from J-CHAT to prompt the model, then performed inference to predict the next 25 seconds of the dialogue based on those initial 5 seconds.
For the sampling method, we used the beam search with a beam size of 5.
When synthesizing speech from discritized speech, we conditioned the vocoder with JVS001 (male) and JVS002 (female) speaker from the JVS corpus~\cite{takamichi2020jsut}. This resulted in the male speaker's voice being heard from the first audio channel (left) and the female speaker's voice from the second channel (right).

\subsection{Result}
Table~\ref{tab:result} shows the results of the subjective evaluation. From these results, it can be seen that dGSLM-J-CHAT achieves the best performance among the dGSLM generated samples in both naturalness and meaningfulness. This suggests that J-CHAT is a useful corpus for constructing generative dialogue language models. We can also see that there is no statistical significance between dGSLM-YouTube and dGSLM-podcast, despite dGSLM-podcast being trained on a dataset approximately four times larger in number of dialogues. This indicates that simply scaling up the dataset is not enough to enhance dGSLM performance. However, there is a significant difference between resynth and dGSLM-J-CHAT in terms of both naturalness and meaningfulness. The trained model occasionally produces sensible words, but the generated dialogue often lacks coherence. This might be improved with better modeling or by increasing the dataset size.
\section{Conclusion}
In this study, we presented J-CHAT, a large-scale Japanese spoken dialogue corpus, and proposed an automated and language-independent methodology for its construction. The experimental results demonstrated that training on diverse dialogue data from multiple data sources, such as YouTube and podcasts, significantly improves spoken dialogue systems models than collecting from single data source.
Future work includes further improving dialogue modeling techniques to enhance the quality of generated speech.

\section{Ethics statement}
The published data were collected and distributed in Japan and therefore comply with Japanese law. The legality of the collection and distribution process of J-CHAT was reviewed by a third-party Japanese lawyer.
Under Article 30-4 of the Japanese Copyright Act, copyrighted materials may be used without explicit permission when the purpose is limited to ``information analysis.''\footref{copyright} Accordingly, the dataset is released strictly for purposes related to information analysis, and not for entertainment or content redistribution.
To mitigate privacy concerns, we performed anonymization of source metadata. In addition, the original recordings were segmented into dialogue units, preventing reconstruction of the original content for entertainment purposes.
We also implement an opt-out policy, following prior large-scale web corpus practices (e.g. Common Crawl), allowing individuals or rights holders to request removal of their data from the dataset.
\section{Acknowledgements}
This work was supported by AIST KAKUSEI project (FY2023), JST Moonshot JPMJMS2011 and JST FOREST JPMJFR226V.
\section{Bibliographical References}
\bibliographystyle{lrec2026-natbib}
\bibliography{anthology,custom}

\end{document}